\documentclass{article} 
\usepackage{samples/iclr2020_conference}
\iclrfinalcopy 

\usepackage{xcolor}
\definecolor{darkblue}{rgb}{0.0, 0.0, 0.55} 
\definecolor{darkcerulean}{rgb}{0.03, 0.27, 0.49}
\definecolor{darkcandyapplered}{rgb}{0.64, 0.0, 0.0}
\definecolor{darklavender}{rgb}{0.45, 0.31, 0.59}
\definecolor{darkmagenta}{rgb}{0.55, 0.0, 0.55}

\usepackage[utf8]{inputenc}
\usepackage[english]{babel}
\usepackage[T1]{fontenc}
\usepackage{microtype}      
\usepackage{times}

\usepackage{graphicx,adjustbox,subcaption,wrapfig}
\usepackage{multirow,booktabs}
\usepackage{url}
\usepackage[colorlinks=true,
            linkcolor=darkblue,
            citecolor=darkblue,
            filecolor=magenta,
            urlcolor= red,]{hyperref}
\usepackage{bookmark} 

\usepackage{natbib}
\setcitestyle{authoryear,open={[},close={]}}

\newcommand{\tocite}[1]{\textbf{\textcolor{blue}{[#1]}}}

\newcommand{\eg}{\textit{e.g.}}
\newcommand{\ie}{\textit{i.e.}}
\newcommand{\viz}{\textit{viz.}}

\newcommand{\etal}{\emph{et al. }}

\usepackage[ruled,noline,linesnumbered]{algorithm2e}
\IncMargin{1em}
\SetKwInOut{Input}{Input}
\SetKwInOut{Output}{Output}

\usepackage{listings}
\definecolor{codegreen}{rgb}{0,0.6,0}
\definecolor{codegray}{rgb}{0.5,0.5,0.5}
\definecolor{codepurple}{rgb}{0.58,0,0.82}
\definecolor{backcolour}{rgb}{0.95,0.95,0.92}
\lstdefinestyle{mystyle}{
    backgroundcolor=\color{backcolour},   
    commentstyle=\color{codegreen},
    keywordstyle=\color{magenta},
    numberstyle=\tiny\color{codegray},
    stringstyle=\color{codepurple},
    basicstyle=\ttfamily\footnotesize,
    breakatwhitespace=false,         
    breaklines=true,                 
    captionpos=b,                    
    keepspaces=true,                 
    showspaces=false,                
    showstringspaces=false,
    showtabs=false,                  
    tabsize=2
    }
\usepackage{amsmath,amsfonts, mathtools, bm, bbm, nicefrac}
\newcommand{\mcal}[1]{\mathcal{#1}}

\newcommand{\E}{\mathbb{E}}

\def\vzero{{\bm{0}}}

\def\vmu{{\bm{\mu}}}

\def\vm{{\bm{m}}}

\def\vv{{\bm{v}}}
\def\vw{{\bm{w}}}
\def\vx{{\bm{x}}}
\def\vy{{\bm{y}}}
\def\vz{{\bm{z}}}

\def\mI{{\bm{I}}}

\DeclareMathAlphabet{\mathsfit}{\encodingdefault}{\sfdefault}{m}{sl}
\SetMathAlphabet{\mathsfit}{bold}{\encodingdefault}{\sfdefault}{bx}{n}

\def\sR{{\mathbb{R}}}

\newcommand{\ptheta}{p_{\theta}}

\title{MagicProp: Diffusion-based Video Editing via Motion-aware Appearance Propagation}
\author{Hanshu Yan\thanks{Equal contribution.}, ~Jun Hao Liew\footnotemark[1], ~Long Mai, ~Shanchuan Lin \& ~Jiashi Feng\\
ByteDance Inc.\\
\texttt{\{hanshu.yan, junhao.liew, long.mai, peterlin, jshfeng\}@bytedance.com}\\
}

\begin{document}

\maketitle

\begin{figure}[h!]
    \centering
    \includegraphics[width=0.95\textwidth]{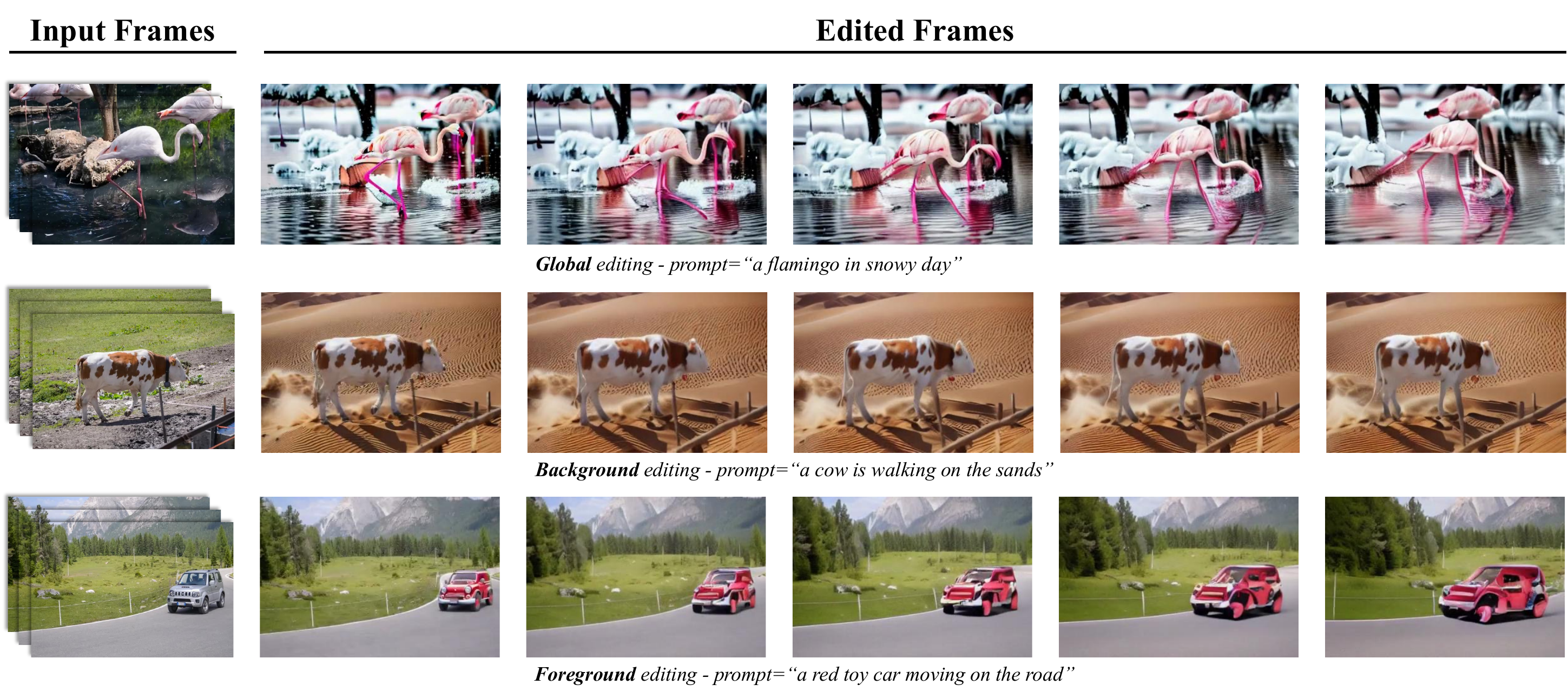}
    \caption{Video editing via MagicProp: global, background, and foreground editing are all supported.}
    \label{fig:teaser}
\end{figure}

\begin{abstract}
    This paper addresses the issue of modifying the visual appearance of videos while preserving their motion. A novel framework, named MagicProp, is proposed, which disentangles the video editing process into two stages: appearance editing and motion-aware appearance propagation. In the first stage, MagicProp selects a single frame from the input video and applies image-editing techniques to modify the content and/or style of the frame. The flexibility of these techniques enables the editing of arbitrary regions within the frame. In the second stage, MagicProp employs the edited frame as an appearance reference and generates the remaining frames using an autoregressive rendering approach. To achieve this, a diffusion-based conditional generation model, called PropDPM, is developed, which synthesizes the target frame by conditioning on the reference appearance, the target motion, and its previous appearance. The autoregressive editing approach ensures temporal consistency in the resulting videos. Overall, MagicProp combines the flexibility of image-editing techniques with the superior temporal consistency of autoregressive modeling, enabling flexible editing of object types and aesthetic styles in arbitrary regions of input videos while maintaining good temporal consistency across frames. Extensive experiments in various video editing scenarios demonstrate the effectiveness of MagicProp.
\end{abstract}

\section{Introduction}
\par Content creation often involves video editing, which includes modifying the appearance or adjusting the motion of raw videos \citep{wu_tune--video_2023, kasten_layered_2021, zhao_controlvideo_2023, wang_videocomposer_2023}. Filmmakers may need to adjust the exposure, saturation, and contrast of raw videos for better aesthetic quality, while advertisers may want to change realistic videos into certain fascinating styles to impress target audiences. This paper addresses the problem of editing videos' appearance, including changing the content or style locally in a certain spatial region or globally throughout the entire video.

\par Existing works attempt to solve this problem mainly from two perspectives: editing each frame individually via image generation models \citep{qi_fatezero_2023, ceylan_pix2video_2023, yang_rerender_2023, khachatryan_text2video-zero_2023, geyer_tokenflow_2023} or modeling the entire video sequence for appearance changing \citep{ni_conditional_2023, molad_dreamix_2023, karras_dreampose_2023, kasten_layered_2021, esser_structure_2023}. Methods based on image models, such as Stable Diffusion \citep{rombach_high-resolution_2022} and ControlNet \citep{zhang_adding_2023}, can flexibly modify the content or style of any arbitrary region, but it is challenging to ensure temporal consistency across adjacent frames. To alleviate this issue, some use structure-guided models and cross-frame attention to align color and layout across frames \citep{zhang_adding_2023, qi_fatezero_2023, ceylan_pix2video_2023}. Other methods exploit inter-frame correspondence, such as optical flow, to warp the features of edited frames \citep{yang_rerender_2023, geyer_tokenflow_2023}. However, the temporal consistency of the edited video is still suboptimal. Instead of using image-based models, researchers have developed many sequence-based models for video generation and editing \citep{esser_structure_2023, couairon_videdit_2023}. Neural Layered Atlas (NLA) overfits a video first and then edits the learned corresponding Atlas to change the foreground or background \citep{kasten_layered_2021, bar-tal_text2live_2022}. NLA-based methods can effectively edit the appearance of videos, but test-time optimization is time- and resource-consuming. Recently, many diffusion-based models have been proposed for structure-aware video generation, such as Gen-1 \citep{esser_structure_2023}, ControlVideo \citep{zhao_controlvideo_2023, chen_control--video_2023}, and VideoComposer \citep{wang_videocomposer_2023}. These methods synthesize videos by conditioning on layout sequences such as depth or sketch maps, so that the motion coherence in the resultant video can be ensured. However, the editability and flexibility will be compromised due to the limitation of textual descriptions and the difficulty of user interaction. For instance, when editing a certain part of a given video, text prompts may not precisely localize the region of interest across all frames, and it may be challenging for users to prepare masks for all frames. The trade-off between temporal consistency and editing flexibility inspires us to explore other alternative frameworks for video editing.

Motivated by the fact that frames within a video usually share a similar scene, we propose a novel framework, MagicProp, which disentangles video editing into two stages, namely, appearance editing and motion-aware appearance propagation. MagicProp first selects one frame from the given video and edits its appearance. The edited frame is used as the appearance reference in the second stage. Then, MagicProp autoregressively renders the remaining frames by conditioning on the reference frame and the motion sequence (\eg, depth maps of the given video). MagicProp models videos in an autoregressive manner, which guarantees the temporal consistency of the output videos. Additionally, MagicProp uses powerful image diffusion models (optionally with additional masks) for reference editing, allowing for flexible modification of the contents of a local region or the entire video.

The most crucial component of MagicProp is an autoregressive conditional image diffusion model that synthesizes the target image under the control of its previous frame, the target depth, and the reference appearance. We design a lightweight adapter to merge and inject the semantic-level and pixel-level information of the reference frame into the image generation process, ensuring that the appearance of the resultant frames aligns well with the reference. During training, 
we follow the strategy of zero terminal signal-to-noise ratio (SNR) \citep{lin_common_2023}, which bridges the gap between the noise schedules during training and inference, resulting in better matching of the color and style of generated frames with the reference. We conducted extensive experiments in several video editing scenarios, including local object/background editing and global stylization. The results demonstrate the effectiveness and flexibility of MagicProp. The contributions of MagicProp are three-fold:
\begin{itemize}
    \item We proposed a novel framework, MagicProp, that decouples video editing into appearance editing and motion-aware appearance propagation.
    \item We devised a lightweight adapter to inject class- and pixel-level features into the diffusion model. We also applied the zero-terminal SNR strategy for training. These techniques facilitate the alignment of the appearance.
    \item Extensive experiments demonstrate that MagicProp can flexibly edit any arbitrary region of the given video and generate high-quality results.
\end{itemize}

\section{Related Works and Preliminaries}
In this section, we first review recent related works on the appearance editing of videos. We categorize them into two groups, \ie, editing a video frame by frame via image models, and modeling the whole frame sequence for editing. Then, we introduce the preliminaries about diffusion probabilistic models and the notation for video editing.

\subsection{Related Works}
\paragraph{Frame-by-frame Editing} 
Diffusion-based image generation models have achieved great success in image generation and editing tasks~\citep{ho_denoising_2020, ho_video_2022, rombach_high-resolution_2022, blattmann_align_2023}. The simplest method for video editing is to edit each frame individually~\citep{meng_sdedit_2022, liew_magicmix_2022, hertz_prompt--prompt_2022}. Although it is flexible to edit each frame and the resultant frames have a good aesthetic quality, the temporal consistency of the whole video is usually inferior. Some methods use the layout condition generation method to edit each frame~\citep{zhang_adding_2023, huang_composer_2023}. For example, ControlNet~\citep{zhang_adding_2023} synthesizes images with the conditioning of a text description and an additional layout map, such as a depth map or an edge map, thus the spatial layout of the edited frame matches that of the original frame. Whilst these methods can guarantee the layout consistency of the edited videos, the appearance of frames (\eg, identity, texture, and color) still changes apparently across frames. To alleviate the issue of temporal consistency, a line of methods rely on cross-frame attention to fuse the latents of edited frames and those of their previous frames (or other reference frames) \citep{qi_fatezero_2023, hertz_prompt--prompt_2022, khachatryan_text2video-zero_2023, ceylan_pix2video_2023}, so that the consistency of shape and style can be improved. Another line of methods exploit the correspondence between frames in the original video and use it to warp the latent or attention maps when generating future frames~\citep{yang_rerender_2023, geyer_tokenflow_2023}. Correspondence-based wrapping may fail due to the occlusion in consecutive frames. In general, methods based on per-frame editing still suffer from temporal consistency across frames.

\paragraph{Editing via Sequential Modeling} Videos are naturally sequential data, and therefore using sequential models for video generation and editing intrinsically benefits temporal consistency. Neural Layered Atlas (NLA)~\citep{kasten_layered_2021, bar-tal_text2live_2022, huang_inve_2023} represents a video through several 2D maps and 2D-to-color atlases. The appearance of objects and backgrounds can be easily edited by modifying the corresponding atlases. However, NLA needs to perform test-time optimization for each video to learn its representations, which is very time-consuming. Recently, diffusion models have been proven effective in modeling sequential data like videos. Many methods use video diffusion models or flatten image diffusion models into video models for video editing~\citep{ho_video_2022, blattmann_align_2023, zhou_magicvideo_2023, wang_videocomposer_2023}. Dreamix~\citep{molad_dreamix_2023} and Tune-A-Video~\citep{wu_tune--video_2023}, fine-tune the video model on the provided video first and then generate a new video by conditioning the textual prompt of the editing instruction. Fine-tuning on the given video cannot sufficiently guarantee that the motion (layout sequence) in the edited video aligns well with the original. To ameliorate this issue, motion-conditioned video diffusion models have been proposed, including Gen-1~\citep{esser_structure_2023}, ControlVideo~\citep{zhao_controlvideo_2023, chen_control--video_2023}, and VideoComposer~\citep{wang_videocomposer_2023}. These methods generate video with the condition of a layout sequence, such as depth or edge maps. When editing, one can extract the layout sequence from the given video first and then generate a new video by conditioning the layout sequence and an editing text prompt. Overall, editing methods based on video models can effectively synthesize temporally consistent videos, but their editability and image quality are not as good as the image-based models at the current stage due to the limitation of textual description and the difficulty of training a good video model. Textual prompts only can provide a high-level semantic description of the desired appearance. It is challenging to locate a specific local editing region of a video based on textual prompts. 

In contrast, MagicProp disentangles appearance editing and appearance propagation. It can flexibly edit the appearance based on powerful image editing methods that can incorporate textural descriptions and localization masks. Besides, synthesizing future frames with an autoregressive model also ensures temporal consistency across frames.

\subsection{Preliminaries}
\paragraph{Denoising Diffusion Probabilistic Model} 
Denoising diffusion probabilistic models (DDPM) are a family of latent generative models that approximate the probability density of training data by reversing the Markovian Gaussian diffusion processes \citep{sohl-dickstein_deep_nodate, ho_denoising_2020}. Concerning a distribution $ q(\vx)$, DDPM models the probability density $ q(\vx)$ as the marginal of the joint distribution between $\vx$ and a series of latent variables $x_{1:T}$, \ie, $p_{\theta}(\vx)=\int p_{\theta}(\vx_{0:T}) d \vx_{1:T}$ with $\quad \vx = \vx_0.$
The joint distribution is defined as a Markov chain with learned Gaussian transitions starting from the standard normal distribution, \ie,
\begin{align}
p_{\theta}(\vx_T)=\mcal{N}(\vx_T; \vzero, \mI) \\
p_{\theta}(\vx_{t-1}|\vx_{t}) = \mcal N(\vx_{t-1}; \vmu_{\theta}(\vx_t, t), \Sigma_{\theta}(\vx_t, t)) \label{eq:appx_gaus_trans}
\end{align}

To perform likelihood maximization of the parameterized marginal $\ptheta(\cdot)$, DDPM uses a fixed Markov Gaussian diffusion process, $q(\vx_{1:T}|\vx_0)$, to approximate the posterior $\ptheta(\vx_{1:T}|\vx_0)$. In specific, two series, $\alpha_{0:T}$ and $\sigma^2_{0:T}$, are defined, where $1=\alpha_0 > \alpha_1 > \dots, >\alpha_T \geq0$ and $0=\sigma^2_0 < \sigma^2_1 < \dots < \sigma^2_T$. For any $t>s\geq0$, $q(\vx_{t}|\vx_{s})=\mathcal{N}(\vx_{t}; \alpha_{t|s}\vx_{s}, \sigma^2_{t|s}\mI),$ where $\alpha_{t|s} = {\alpha_t}/{\alpha_s}$ and $\sigma^2_{t|s} = \sigma^2_{t} - \alpha^2_{t|s} \sigma^2_{s}.$ Usually, we set $\alpha_t^2 + \sigma_t^2=1$, thus, 
\begin{equation}
    q_(\vx_t|\vx_0) = \mathcal{N}(\vx_t|\alpha_t \vx_0, (1-\alpha_t^2) \mI).
    \label{eq:gt_posterior}
\end{equation}

We use deep neural networks to parameterize the expectation function $\mu_{\theta}(\vx_t, t)$ of the sampling process or the denoising function $\epsilon_{\theta}(\vx_t, t)$, which can be used to alternatively estimate the expectation via $\mu_{\theta}(\vx_t, t) = \frac{1}{\sqrt{\alpha_{t|t-1}}}(\vx_t - \frac{1-\alpha_{t|t-1}}{\sqrt{1-\alpha_t}}\epsilon_{\theta}(\vx_t, t)).$ When performing conditional generation tasks, the network should take additional control signals $\vy$ as input, \ie, $\epsilon_{\theta}(\vx_t, t, \vy)$. 
The parameterized reversed process $\ptheta$ can be optimized by maximizing the associated evidence lower bound (ELBO). We plug the Gaussian parameterization into KL-divergence terms, the ELBO optimization turns to be noise estimation, where $\lambda(t)$ is a weighting function. After training, we can sample new data via the Markov chain defined in Eqn~\eqref{eq:appx_gaus_trans}. Instead, we also can use deterministic samplers, such as DDIM, to generate new data. For a certain starting noise $\vx_T \sim \mathcal{N}(\vx_T; \vzero, \mI)$, the mapping from $\vx_T$ to the generated datum $\vx_0$ through a deterministic sampler is denoted by $\Phi(\vx_T, \vy)$.
\begin{align}
    L = \E_{\vx_0, t, \epsilon}[\lambda(t) \|\epsilon_{\theta}(\vx_t) - \epsilon \|^2_2].
    \label{eq:loss_ddpm}
\end{align}

\paragraph{Notation for Video Editing} 
We denote a video by $\vx=[\vx^1, ..., \vx^K]$, where $\vx^i$ represents the $i^{\text{th}}$ frame in the sequence and, for each $i \in [1,\dots,K]$, $\vx^i \in [-1,1]^{C\times H\times W}$. To reduce the computational overhead of modeling videos, we use a variational auto-encoder (VAE), denoted by $\{\mathcal{E}(\cdot), \mathcal{D}(\cdot)\}$, to map videos from the RGB space to a lower-dimensional latent space. The video frames are transformed one by one, \ie, $\vz=[\vz^1, ..., \vz^K]$ with $\vz^i = \mathcal{E}(\vx^i)$. We follow Stable Diffusion which uses an encoder to downsample $\vx$ into a spatially $8\times$ smaller space. The generated latent codes can be decoded to videos by $\mathcal{D}(\cdot)$. The editing operations require users to provide extra information describing the desired appearance of the target video. We denote the instruction information by $\vy$; it could be a textual description, an extra localization mask, or other visual reference. We use CLIP, denoted by $\tau(\cdot)$, to encode the text prompt or reference image, and the embedding is denoted $\tau(\vy)$. To preserve the motion of the original video, we use a depth estimation model, such as TCMonoDepth, to extract the sequence of depth maps for representing the motion. We denote $\mathcal{M}(\cdot)$ as the depth model and  $\vm=[\vm^1, \dots, \vm^K]$ with $\vm^i = \mathcal{M}(\vx^1)$ as the depth sequence.

\section{Method}

This paper addresses the problem of motion-preserving video editing, where we aim to alter the appearance of a given video while retaining the original motion. Typically, frames in a short video have similar scenes, with main objects and backgrounds appearing consistently throughout. It is natural to disentangle the video editing problem into two sub-tasks, \viz, editing the appearance of the main objects and/or the background first and then propagating the edited content to all other frames based on the original motion.

In this section, we elucidate the pipeline of MagicProp $\mathcal{V}(\cdot)$, 
which performs video editing in two stages sequentially, \ie, appearance editing $\Phi^1(\cdot)$ and motion-aware appearance propagation $\Phi^2(\cdot)$. MagicProp can flexibly edit the appearance of a given video according to users' instructions. It supports changing the contents (\eg, object type and image style) in any specific region, either locally or globally. Formally, MagicProp takes input as the source video $\vx$, a textual prompt $\vy$, and optionally a localization mask $\vw$. This mask can be provided by users or easily obtained by a powerful segmentation model. After the two-stage processing, MagicProp generates an edited video $\hat{\vx}$ whose motion remains unchanged.

\begin{figure}[t!]
    \centering
    \includegraphics[width=0.95\textwidth]{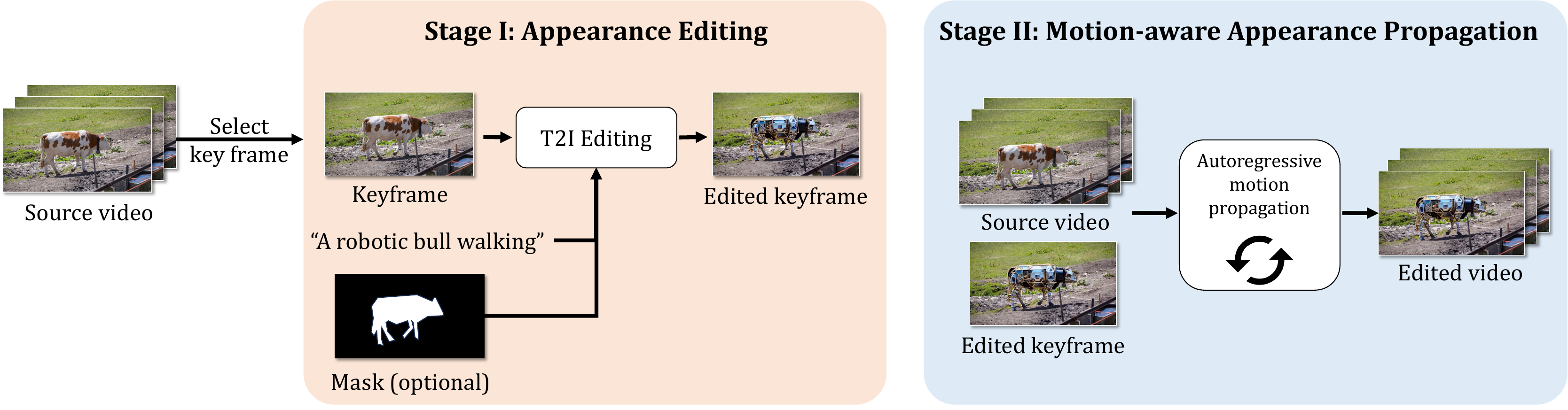}
    \caption{The pipeline of MagicProp.}
    \label{fig:magicprop}
\end{figure}

\subsection{Appearance Editing}
The first stage of MagicProp is to manipulate the appearance of the source video. We select one frame as the appearance reference. Thanks to many effective image-editing methods, we can flexibly edit any arbitrary region of the reference frame, including changing object types or visual styles.

In specific, we select a frame $\vx^{\#}$ from the input video $\vx$ as the appearance reference. Existing image editing methods, such as Text-to-Image (T2I) models, offer rich possibilities to manipulate images' contents~\citep{meng_sdedit_2022, liew_magicmix_2022, zhang_adding_2023}. Here, we use the ControlNet optionally with a segmentation mask $\vw$ to change the main objects and/or the background. By conditioning the depth map of $\vx^{\#}$ and a textual prompt $\vy$, ControlNet will generate a new image $\hat{\vx}^{\#}$ whose layout matches the original one and semantics aligns with the text description. In comparison to existing Text-to-Video (T2V) models, T2I models, such as Stale Diffusion, have apparent superiority in terms of per-frame quality. Thus, the resultant frame edited by ControlNet contains rich details and enjoys high aesthetic quality. Besides, T2I diffusion models allow us to use localization masks to precisely control the editing parts in images. It is flexible to edit a local region or the whole image. In brief, stage one chooses and edits a certain frame, and the edited frame will be used as the appearance reference for video synthesis in the second stage.
\begin{align}
    \hat{\vx}^{\#} = \Phi^1(\vx, \#, \vy, \vw)  
\end{align}

\subsection{Motion-aware Appearance Propagation}

\begin{figure}[t!]
    \centering
    \includegraphics[width=0.95\textwidth]{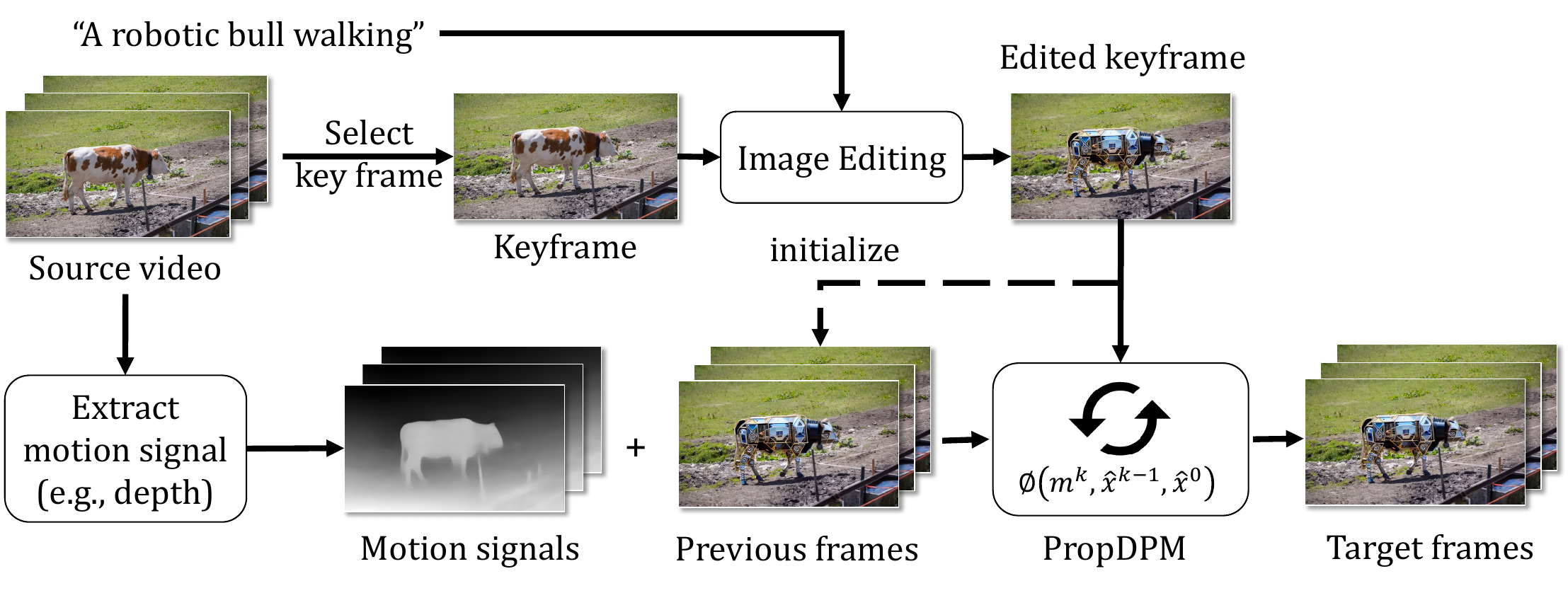}
    \caption{Auto-regressive Motion-aware Appearance Propagation Diffusion Model}
    \label{fig:prop-dpm}
\end{figure}

Given a source video $\vx$ and the appearance reference $\hat{\vx}^{\#}$, the second stage $\Phi^2(\cdot)$ will render a new video $\hat{\vx}$ that preserves the motion in source one and whose appearance matches the reference. The most crucial part is an appearance propagation diffusion probabilistic model (PropDPM). PropDPM, denoted by $\phi_{\theta}(\cdot)$, synthesizes the whole video in an auto-regressive manner. Each frame $\hat{\vx}^k$ is generated with the conditioning of the reference appearance $\hat{\vx}^{\#}$, its corresponding depth map $\vm^k$, and the previous edited frame $\hat{\vx}^{k-1}$. 
We can use the edited appearance reference as the starting frame, \ie,
$\hat{x}^{0} = \hat{x}^{\#}$ and $\vm^{0}=\vm^{\#}$. The rest can be rendered frame-by-frame through Eqn~\eqref{eq:prop-dpm} for $k$ from $1$ to $K$.
The layout in the generated frames aligns with the depth maps extracted from the corresponding frames in the source video. Hence, the motion (layout sequence) remains unchanged compared to the source video, and the temporal consistency in the rendered video is also guaranteed. 
\begin{align}
    \hat{\vx}^k = \phi_{\theta}(\vm^k, \hat{\vx}^{k-1}, \vm^{k-1}, \hat{\vx}^{\#})
    \label{eq:prop-dpm} \\
    \hat{\vx} = \Phi^2(\hat{x}^{\#}, \vx)
\end{align}

In specific, PropDPM is designed based on the latent diffusion model~\citep{rombach_high-resolution_2022}. We use a VAE $\{\mathcal{E}(\cdot), \mathcal{D}(\cdot)\}$ to map a video into a lower-dimensional latent space. PropDPM is trained to generate the edited latent $\hat{\vz}^k$ and we then use the VAE to reconstruct the edited video frame $\hat{\vx}^k$.
For the conditioning signals, we split them into two groups, \viz, the spatial conditions and the semantic conditions. The spatial conditions, including the target frame's depth map and the previous frame, provide the spatial layout information for the generated image and form a contrast between two consecutive frames. This contrast facilitates the synthesis of contents by querying spatially corresponding regions. The semantic conditions include the RGB and the latent of the reference frame. They provide information about the color, style, and object classes in the target edited video. 

The spatial conditions are injected into the PropDPM by concatenating them to the noisy latent. We use the TCMonoDepth~\citep{li_enforcing_tcmonodepth_2021} model to estimate depth maps in the RGB space and rescale them into the size of the latent codes. When generating the $k^{\text{th}}$ edited frame, we concatenate its depth map $\vm^{k}$, the latent of the previous edited frame $\hat{\vz}^{k-1}_t$, the previous depth map $\vm^{k-1}$, to the noisy latent $\hat{\vz}_t$. Instead, the semantic conditions are used as the input of the cross-attention modules. We design a lightweight adaptor to combine the CLIP's embedding and the VAE latent of the reference frame so that the injected semantics contains both class-wise and patch-wise information.

\subsection{Model Design of PropDPM}
The main challenges of video editing are ensuring temporal consistency across all frames and maintaining per-frame quality. PropDPM addresses the first challenge by editing a video in an auto-regressive manner, conditioning on the true depth sequence to ensure temporal coherence across frames. However, due to the intrinsic error accumulation issue of auto-regressive modeling, the image quality of the edited frames degrades as the frame index increases. While the early edited frames contain rich details, the later edited ones become smooth and suffer from color shifting.

To alleviate the error accumulation issue, we propose two complementary solutions. First, we design an appearance adaptor that merges the class-level and patch-wise information of the reference frame. The output of this adaptor is sent to cross-attention modules. During inference, we use a fixed reference frame for each video when auto-regressively synthesizing frames. A fixed reference frame serves as an anchor to ameliorate the degradation. Second, we apply the Zero-Terminal-SNR~\citep{lin_common_2023} technique to train the diffusion model, which bridges the gap between the starting noise's strength during inference and the largest noise level during training. This technique improves the image quality of the generated frame in each iteration.

\subsubsection{Appearance Adaptor}
We design a lightweight adaptor to fuse the class-level and pixel-level features of the reference frame. The adaptor preserves the spatial correspondence between the fused tokens and the reference image. In detail, we first use the VAE to extract the latent of the reference image, $\vz^{\#} \in \sR^{4\times h \times w}$. The latent codes of VAE have good spatial correspondence to the original images. We use a nonlinear network to decrease the redundant spatial resolution of latent $\vz^{\#}$ by a factor of $\times 2$ but increase the channel dimension to preserve more information. The resultant feature is in size of $\sR^{l/2 \times h/2 \times w/2}$, where $l$ is the length of each CLIP embedding. On the other hand, we use the CLIP model to extract the semantics of the reference image. We have a global class token $\tau(\vx^{\#})_{\text{c}} \in \sR^{l \times 1}$ and patch-wise tokens $\tau(\vx^{\#})_{\text{p}} \in \sR^{l \times h'\times w'}$. We utilize another nonlinear network to downsample the token dimension and adjust their spatial resolution to $\sR^{l/2 \times h/2 \times w/2}$. Finally, we apply the third nonlinear module to merge the transformed CLIP's and the VAE's features into a fused feature in size of $\sR^{l \times h/2 \times w/2}$. We concatenate it with the untouched class-level token and use it (reshaped into the size of $\sR^{l \times (hw/4+1)}$) as the input of cross-attention modules. Since the fused tokens contain rich global and local information, PropDPM can generate a target frame that better matches the reference's appearance.


\subsubsection{Zero-Terminal-SNR Noise Schedule}
Diffusion models are trained to estimate the noise in the noisy intermediate state $\vx_t$ for $t\in [1, \dots, T]$,  where $\vx_t = \alpha_t \vx_0 + \sqrt{1-\alpha^2_t} \epsilon$. In the vanilla DDPM, the noise schedule is set to be $1=\alpha_0 > \alpha_1 > \dots > \alpha_T > 0$, where the terminal signal-to-noise-ratio (SNR), $SNR(t)=\alpha_t^2 / (1-\alpha_t^2)$, is greater than 0. This means the strongest noise, that the obtained DDPM can handle, is $\vx_T = \alpha_T \vx_0 + \sqrt{1-\alpha^2_T} \epsilon$ rather than the pure noise $\epsilon$. However, during inference, most samplers start from pure noise. This gap may incur the degradation of the generated data. To fix this issue, Lin \etal \tocite{ } propose a novel noise schedule, termed Zero-Terminal-SNR, which forces the $\text{SNR}(T)$ to be zero and make the UNet $\vv_{\theta}(\vz_t)$ to predict the $\vv$-value instead of noise $\epsilon$. The $v$-value is defined as $\vv_t = \alpha_t \epsilon - \sqrt{(1-\alpha_t^2)} \vx_0$. 
We follow the Zero-Terminal-SNR strategy for training our PropDPM model. The experimental results verify the effectiveness of alleviating the color-shifting issue.

\subsubsection{Training}
The PropDPM is initialized from the Stable-Diffusion-v1.5. We train the PropDPM model on the combination of a public video dataset, WebVid-10M~\citep{Bain21}, and a self-collected private dataset. For the public one, we randomly sample 1 million videos, while the self-collected dataset contains 20 thousand high-resolution videos without watermarks. From each video, we sample at most 30 frames with a step size of \textit{four}. These frames are then center-cropped into squares and resized into the shape of $256\times 256$. During training, we randomly select three frames from a video to form a triplet: the reference frame, the previous frame, and the target frame.

\section{Application}

\begin{figure}[h!]
    \centering
    \includegraphics[width=0.95\textwidth]{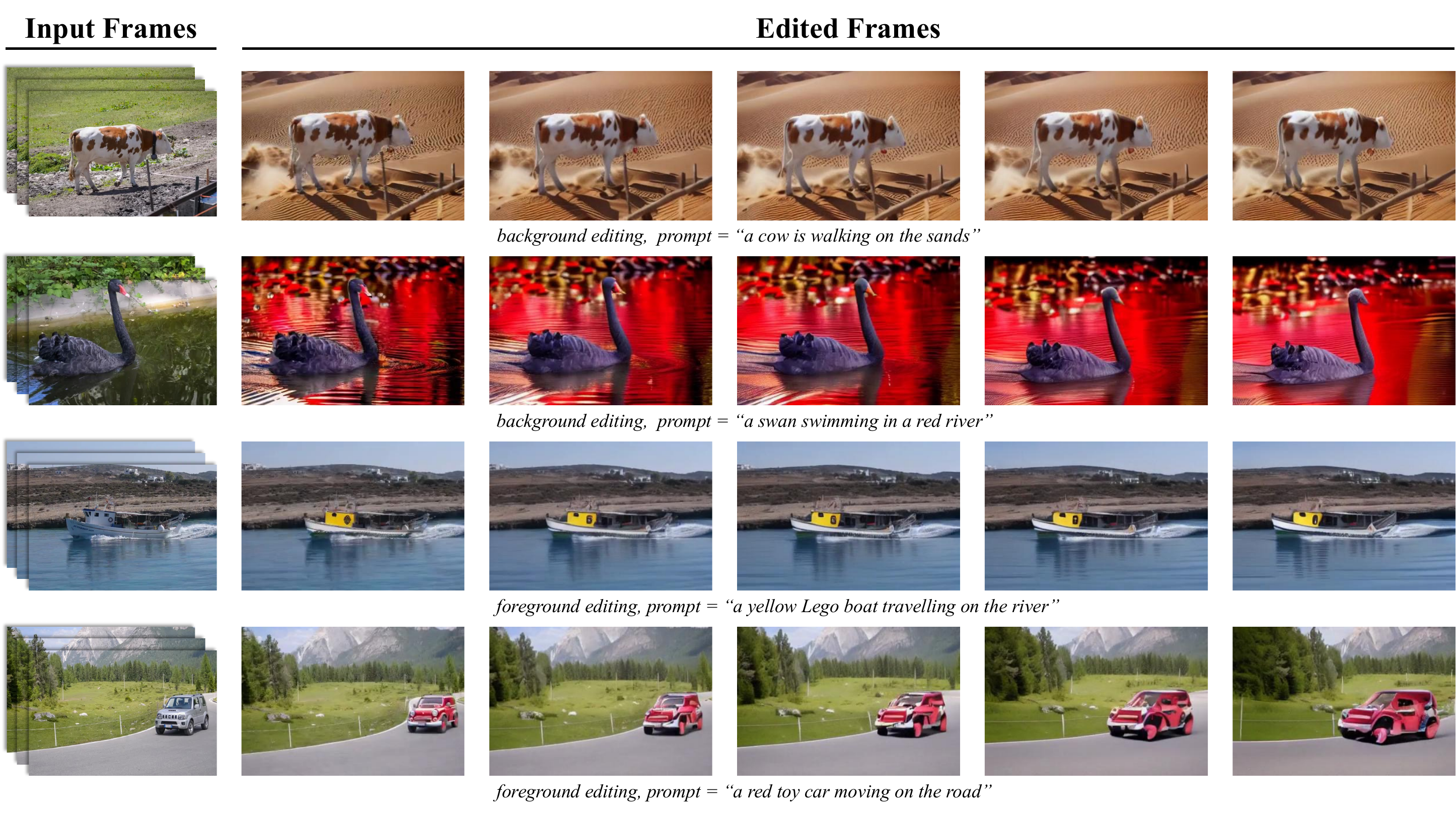}
    \caption{Examples for local editing---background (the top two) and foreground editing (the bottom two).}
    \label{fig:local-editing}
\end{figure}

\begin{figure}[h!]
    \centering
    \includegraphics[width=0.95\textwidth]{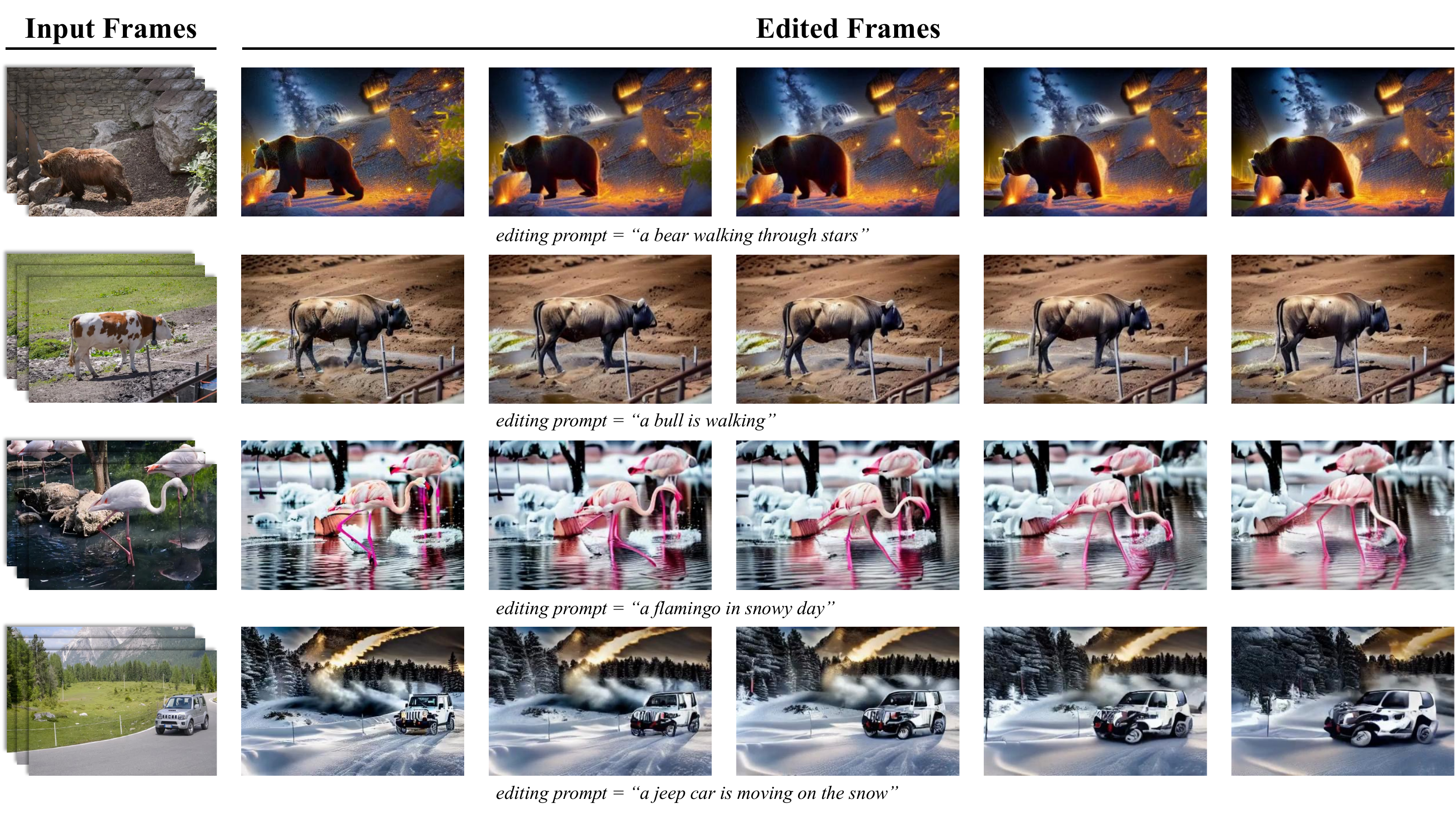}
    \caption{Examples for global editing.}
    \label{fig:global-editing}
\end{figure}

MagicProp can edit any arbitrary region in the given video. In Figure~\ref{fig:local-editing} and Figure~\ref{fig:global-editing}, we show the rendered videos. We use masks and ControlNet to localize and modify certain parts. The masks can be either provided by users or extracted by a segmentation model (\eg, Segment-Anything). 

Through extensive experiments, we find MagicProp can robustly edit videos up to 30 frames. Degradation, such as over-smoothing and artifacts, may appear when the length of videos exceeds 30 frames due to the intrinsic error accumulation of Auto-regressive inference. For future work, we aim to improve the current MagicProp framework for processing longer videos. 

\bibliography{samples/_references}

\begin{thebibliography}{31}
\providecommand{\natexlab}[1]{#1}
\providecommand{\url}[1]{\texttt{#1}}
\expandafter\ifx\csname urlstyle\endcsname\relax
  \providecommand{\doi}[1]{doi: #1}\else
  \providecommand{\doi}{doi: \begingroup \urlstyle{rm}\Url}\fi

\bibitem[Bain et~al.(2021)Bain, Nagrani, Varol, and Zisserman]{Bain21}
M.~Bain, A.~Nagrani, G.~Varol, and A.~Zisserman.
\newblock Frozen in time: A joint video and image encoder for end-to-end
  retrieval.
\newblock In \emph{IEEE International Conference on Computer Vision}, 2021.

\bibitem[Bar-Tal et~al.(2022)Bar-Tal, Ofri-Amar, Fridman, Kasten, and
  Dekel]{bar-tal_text2live_2022}
O.~Bar-Tal, D.~Ofri-Amar, R.~Fridman, Y.~Kasten, and T.~Dekel.
\newblock {Text2LIVE}: {Text}-{Driven} {Layered} {Image} and {Video} {Editing},
  May 2022.
\newblock arXiv:2204.02491 [cs].

\bibitem[Blattmann et~al.(2023)Blattmann, Rombach, Ling, Dockhorn, Kim, Fidler,
  and Kreis]{blattmann_align_2023}
A.~Blattmann, R.~Rombach, H.~Ling, T.~Dockhorn, S.~W. Kim, S.~Fidler, and
  K.~Kreis.
\newblock Align your {Latents}: {High}-{Resolution} {Video} {Synthesis} with
  {Latent} {Diffusion} {Models}, Apr. 2023.
\newblock arXiv:2304.08818 [cs].

\bibitem[Ceylan et~al.(2023)Ceylan, Huang, and Mitra]{ceylan_pix2video_2023}
D.~Ceylan, C.-H.~P. Huang, and N.~J. Mitra.
\newblock {Pix2Video}: {Video} {Editing} using {Image} {Diffusion}, Mar. 2023.
\newblock arXiv:2303.12688 [cs].

\bibitem[Chen et~al.(2023)Chen, Wu, Xie, Wu, Li, Xia, Xiao, and
  Lin]{chen_control--video_2023}
W.~Chen, J.~Wu, P.~Xie, H.~Wu, J.~Li, X.~Xia, X.~Xiao, and L.~Lin.
\newblock Control-{A}-{Video}: {Controllable} {Text}-to-{Video} {Generation}
  with {Diffusion} {Models}, May 2023.
\newblock arXiv:2305.13840 [cs].

\bibitem[Couairon et~al.(2023)Couairon, Rambour, Haugeard, and
  Thome]{couairon_videdit_2023}
P.~Couairon, C.~Rambour, J.-E. Haugeard, and N.~Thome.
\newblock {VidEdit}: {Zero}-{Shot} and {Spatially} {Aware} {Text}-{Driven}
  {Video} {Editing}, June 2023.
\newblock arXiv:2306.08707 [cs].

\bibitem[Esser et~al.(2023)Esser, Chiu, Atighehchian, Granskog, and
  Germanidis]{esser_structure_2023}
P.~Esser, J.~Chiu, P.~Atighehchian, J.~Granskog, and A.~Germanidis.
\newblock Structure and {Content}-{Guided} {Video} {Synthesis} with {Diffusion}
  {Models}, Feb. 2023.
\newblock arXiv:2302.03011 [cs].

\bibitem[Geyer et~al.(2023)Geyer, Bar-Tal, Bagon, and
  Dekel]{geyer_tokenflow_2023}
M.~Geyer, O.~Bar-Tal, S.~Bagon, and T.~Dekel.
\newblock {TokenFlow}: {Consistent} {Diffusion} {Features} for {Consistent}
  {Video} {Editing}, July 2023.
\newblock arXiv:2307.10373 [cs].

\bibitem[Hertz et~al.(2022)Hertz, Mokady, Tenenbaum, Aberman, Pritch, and
  Cohen-Or]{hertz_prompt--prompt_2022}
A.~Hertz, R.~Mokady, J.~Tenenbaum, K.~Aberman, Y.~Pritch, and D.~Cohen-Or.
\newblock Prompt-to-{Prompt} {Image} {Editing} with {Cross} {Attention}
  {Control}, Aug. 2022.
\newblock arXiv:2208.01626 [cs].

\bibitem[Ho et~al.(2020)Ho, Jain, and Abbeel]{ho_denoising_2020}
J.~Ho, A.~Jain, and P.~Abbeel.
\newblock Denoising {Diffusion} {Probabilistic} {Models}, Dec. 2020.
\newblock arXiv:2006.11239 [cs, stat].

\bibitem[Ho et~al.(2022)Ho, Salimans, Gritsenko, Chan, Norouzi, and
  Fleet]{ho_video_2022}
J.~Ho, T.~Salimans, A.~Gritsenko, W.~Chan, M.~Norouzi, and D.~J. Fleet.
\newblock Video {Diffusion} {Models}.
\newblock Technical Report arXiv:2204.03458, arXiv, June 2022.
\newblock arXiv:2204.03458 [cs] type: article.

\bibitem[Huang et~al.(2023{\natexlab{a}})Huang, Sigal, Yi, Wang, and
  Lee]{huang_inve_2023}
J.~Huang, L.~Sigal, K.~M. Yi, O.~Wang, and J.-Y. Lee.
\newblock {INVE}: {Interactive} {Neural} {Video} {Editing}, July
  2023{\natexlab{a}}.
\newblock arXiv:2307.07663 [cs].

\bibitem[Huang et~al.(2023{\natexlab{b}})Huang, Chen, Liu, Shen, Zhao, and
  Zhou]{huang_composer_2023}
L.~Huang, D.~Chen, Y.~Liu, Y.~Shen, D.~Zhao, and J.~Zhou.
\newblock Composer: {Creative} and {Controllable} {Image} {Synthesis} with
  {Composable} {Conditions}, Feb. 2023{\natexlab{b}}.
\newblock arXiv:2302.09778 [cs].

\bibitem[Karras et~al.(2023)Karras, Holynski, Wang, and
  Kemelmacher-Shlizerman]{karras_dreampose_2023}
J.~Karras, A.~Holynski, T.-C. Wang, and I.~Kemelmacher-Shlizerman.
\newblock {DreamPose}: {Fashion} {Image}-to-{Video} {Synthesis} via {Stable}
  {Diffusion}, May 2023.
\newblock arXiv:2304.06025 [cs].

\bibitem[Kasten et~al.(2021)Kasten, Ofri, Wang, and Dekel]{kasten_layered_2021}
Y.~Kasten, D.~Ofri, O.~Wang, and T.~Dekel.
\newblock Layered {Neural} {Atlases} for {Consistent} {Video} {Editing}, Sept.
  2021.
\newblock arXiv:2109.11418 [cs].

\bibitem[Khachatryan et~al.(2023)Khachatryan, Movsisyan, Tadevosyan, Henschel,
  Wang, Navasardyan, and Shi]{khachatryan_text2video-zero_2023}
L.~Khachatryan, A.~Movsisyan, V.~Tadevosyan, R.~Henschel, Z.~Wang,
  S.~Navasardyan, and H.~Shi.
\newblock {Text2Video}-{Zero}: {Text}-to-{Image} {Diffusion} {Models} are
  {Zero}-{Shot} {Video} {Generators}, Mar. 2023.
\newblock arXiv:2303.13439 [cs].

\bibitem[Li et~al.(2021)Li, Luo, Zhu, Zhao, Li, and
  Shan]{li_enforcing_tcmonodepth_2021}
S.~Li, Y.~Luo, Y.~Zhu, X.~Zhao, Y.~Li, and Y.~Shan.
\newblock Enforcing temporal consistency in video depth estimation.
\newblock In \emph{2021 IEEE/CVF International Conference on Computer Vision
  Workshops (ICCVW)}, pages 1145--1154, 2021.
\newblock \doi{10.1109/ICCVW54120.2021.00134}.

\bibitem[Liew et~al.(2022)Liew, Yan, Zhou, and Feng]{liew_magicmix_2022}
J.~H. Liew, H.~Yan, D.~Zhou, and J.~Feng.
\newblock {MagicMix}: {Semantic} {Mixing} with {Diffusion} {Models}, Oct. 2022.
\newblock arXiv:2210.16056 [cs].

\bibitem[Lin et~al.(2023)Lin, Liu, Li, and Yang]{lin_common_2023}
S.~Lin, B.~Liu, J.~Li, and X.~Yang.
\newblock Common {Diffusion} {Noise} {Schedules} and {Sample} {Steps} are
  {Flawed}, May 2023.
\newblock arXiv:2305.08891 [cs].

\bibitem[Meng et~al.(2022)Meng, He, Song, Song, Wu, Zhu, and
  Ermon]{meng_sdedit_2022}
C.~Meng, Y.~He, Y.~Song, J.~Song, J.~Wu, J.-Y. Zhu, and S.~Ermon.
\newblock {SDEdit}: {Guided} {Image} {Synthesis} and {Editing} with
  {Stochastic} {Differential} {Equations}.
\newblock Technical Report arXiv:2108.01073, arXiv, Jan. 2022.
\newblock arXiv:2108.01073 [cs] type: article.

\bibitem[Molad et~al.(2023)Molad, Horwitz, Valevski, Acha, Matias, Pritch,
  Leviathan, and Hoshen]{molad_dreamix_2023}
E.~Molad, E.~Horwitz, D.~Valevski, A.~R. Acha, Y.~Matias, Y.~Pritch,
  Y.~Leviathan, and Y.~Hoshen.
\newblock Dreamix: {Video} {Diffusion} {Models} are {General} {Video}
  {Editors}, Feb. 2023.
\newblock arXiv:2302.01329 [cs].

\bibitem[Ni et~al.(2023)Ni, Shi, Li, Huang, and Min]{ni_conditional_2023}
H.~Ni, C.~Shi, K.~Li, S.~X. Huang, and M.~R. Min.
\newblock Conditional {Image}-to-{Video} {Generation} with {Latent} {Flow}
  {Diffusion} {Models}, Mar. 2023.
\newblock arXiv:2303.13744 [cs].

\bibitem[Qi et~al.(2023)Qi, Cun, Zhang, Lei, Wang, Shan, and
  Chen]{qi_fatezero_2023}
C.~Qi, X.~Cun, Y.~Zhang, C.~Lei, X.~Wang, Y.~Shan, and Q.~Chen.
\newblock {FateZero}: {Fusing} {Attentions} for {Zero}-shot {Text}-based
  {Video} {Editing}, Mar. 2023.
\newblock arXiv:2303.09535 [cs].

\bibitem[Rombach et~al.(2022)Rombach, Blattmann, Lorenz, Esser, and
  Ommer]{rombach_high-resolution_2022}
R.~Rombach, A.~Blattmann, D.~Lorenz, P.~Esser, and B.~Ommer.
\newblock High-{Resolution} {Image} {Synthesis} with {Latent} {Diffusion}
  {Models}, Apr. 2022.
\newblock arXiv:2112.10752 [cs].

\bibitem[Sohl-Dickstein et~al.(2015)Sohl-Dickstein, Weiss, Maheswaranathan, and
  Ganguli]{sohl-dickstein_deep_nodate}
J.~Sohl-Dickstein, E.~A. Weiss, N.~Maheswaranathan, and S.~Ganguli.
\newblock Deep {Unsupervised} {Learning} using {Nonequilibrium}
  {Thermodynamics}.
\newblock page~10, 2015.

\bibitem[Wang et~al.(2023)Wang, Yuan, Zhang, Chen, Wang, Zhang, Shen, Zhao, and
  Zhou]{wang_videocomposer_2023}
X.~Wang, H.~Yuan, S.~Zhang, D.~Chen, J.~Wang, Y.~Zhang, Y.~Shen, D.~Zhao, and
  J.~Zhou.
\newblock {VideoComposer}: {Compositional} {Video} {Synthesis} with {Motion}
  {Controllability}, June 2023.
\newblock arXiv:2306.02018 [cs].

\bibitem[Wu et~al.(2023)Wu, Ge, Wang, Lei, Gu, Shi, Hsu, Shan, Qie, and
  Shou]{wu_tune--video_2023}
J.~Z. Wu, Y.~Ge, X.~Wang, W.~Lei, Y.~Gu, Y.~Shi, W.~Hsu, Y.~Shan, X.~Qie, and
  M.~Z. Shou.
\newblock Tune-{A}-{Video}: {One}-{Shot} {Tuning} of {Image} {Diffusion}
  {Models} for {Text}-to-{Video} {Generation}, Mar. 2023.
\newblock arXiv:2212.11565 [cs].

\bibitem[Yang et~al.(2023)Yang, Zhou, Liu, and Loy]{yang_rerender_2023}
S.~Yang, Y.~Zhou, Z.~Liu, and C.~C. Loy.
\newblock Rerender {A} {Video}: {Zero}-{Shot} {Text}-{Guided}
  {Video}-to-{Video} {Translation}, June 2023.
\newblock arXiv:2306.07954 [cs].

\bibitem[Zhang and Agrawala(2023)]{zhang_adding_2023}
L.~Zhang and M.~Agrawala.
\newblock Adding {Conditional} {Control} to {Text}-to-{Image} {Diffusion}
  {Models}, Feb. 2023.
\newblock arXiv:2302.05543 [cs].

\bibitem[Zhao et~al.(2023)Zhao, Wang, Bao, Li, and Zhu]{zhao_controlvideo_2023}
M.~Zhao, R.~Wang, F.~Bao, C.~Li, and J.~Zhu.
\newblock {ControlVideo}: {Adding} {Conditional} {Control} for {One} {Shot}
  {Text}-to-{Video} {Editing}, May 2023.
\newblock arXiv:2305.17098 [cs].

\bibitem[Zhou et~al.(2023)Zhou, Wang, Yan, Lv, Zhu, and
  Feng]{zhou_magicvideo_2023}
D.~Zhou, W.~Wang, H.~Yan, W.~Lv, Y.~Zhu, and J.~Feng.
\newblock {MagicVideo}: {Efficient} {Video} {Generation} {With} {Latent}
  {Diffusion} {Models}, May 2023.
\newblock arXiv:2211.11018 [cs].

\end{thebibliography}

\appendix

\section{Supplementary}

Implementation of the adaptor that fuses the class-level and pixel-level information of the edited reference frame.
\begin{lstlisting}[language=Python, caption=Appearance adaptor]
import math
import numpy as np
import torch
import torch.nn as nn
from einops import rearrange

class Embedding_Adapter(nn.Module):
    """
        Fusing the CLIP embeddings and VAE latents of images.
    """
    def __init__(self, ca_emb_size=768):
        super(Embedding_Adapter, self).__init__()
        assert ca_emb_size % 2 == 0
        self.clip_downsample_emb = nn.Sequential(
                        nn.Conv2d(ca_emb_size, ca_emb_size//2, 
                            kernel_size=1, stride=1, 
                            padding=0, bias=True),
                        nn.SiLU(),
                        nn.Conv2d(ca_emb_size//2, ca_emb_size//2, 
                            kernel_size=1, stride=1, 
                            padding=0, bias=True),
                        nn.SiLU(),
                        )
        self.vae_upsample_chn_down_spatial = nn.Sequential(
                        nn.Conv2d(4, ca_emb_size//2, 
                            kernel_size=3, stride=1, 
                            padding=1, bias=True),
                        nn.SiLU(),
                        nn.MaxPool2d(2),
                        nn.Conv2d(ca_emb_size//2, ca_emb_size//2, 
                            kernel_size=1, stride=1, 
                            padding=0, bias=True),
                        nn.SiLU(),
                        )
        self.mix_clip_vae = nn.Sequential(
                        nn.Conv2d(ca_emb_size, ca_emb_size, 
                            kernel_size=3, stride=1, 
                            padding=1, bias=True),
                        nn.SiLU(),
                        nn.Conv2d(ca_emb_size, ca_emb_size, 
                            kernel_size=3, stride=1, 
                            padding=1, bias=True),
                        )

    def forward(self, clip, vae):
        bs_vae, _, h_vae, w_vae = vae.size()
        bs_emb, hw, _ = clip.size()
        h_emb = int(math.sqrt(hw-1))
        assert (bs_vae==bs_emb) and ((hw-1)%h_emb==0)
        
        clip_cls = clip[:, 0:1, :]
        
        ## adjusting clip_patch embeddings, 
        clip_patch = clip[:, 1:, :]
        clip_patch = rearrange(clip_patch, 'b (h w) c -> b c h w', 
                                h=h_emb, w=h_emb)
        clip_patch = torch.nn.functional.interpolate(clip_patch, 
                                size=(h_vae//2, w_vae//2), 
                                mode="bilinear")
        # (b, h_emb^2, 768) -> (b, 384, h//2, w//2)
        clip_patch = self.clip_downsample_emb(clip_patch) 
        
        ## adjusting vae latents
        # (b, 4, h, w) -> (b, 384, h//2, w//2)
        vae = self.vae_upsample_chn_down_spatial(vae) 
        
        ## fusing, #(b, 1+hw//4, 768)
        clip_vae = torch.cat([clip_patch, vae], dim=1)
        clip_vae = self.mix_clip_vae(clip_vae)
        clip_vae = rearrange(clip_vae, 'b c h w -> b (h w) c') 
        return torch.cat([clip_cls, clip_vae], dim=1)
\end{lstlisting}

\end{document}